\documentclass{article}

\usepackage{microtype}
\usepackage{graphicx}
\usepackage{booktabs} 

\usepackage{hyperref}

\usepackage[accepted]{icml2025}

\usepackage{amsmath}
\usepackage{amssymb}
\usepackage{mathtools}
\usepackage{amsthm}

\usepackage[capitalize,noabbrev]{cleveref}

\theoremstyle{plain}
\newtheorem{theorem}{Theorem}[section]

\theoremstyle{definition}
\newtheorem{definition}[theorem]{Definition}

\theoremstyle{remark}

\usepackage[textsize=tiny]{todonotes}

\usepackage[utf8]{inputenc} 
\usepackage{url}            
\usepackage{booktabs}       
\usepackage{amsfonts}       
\usepackage{nicefrac}       
\usepackage{microtype}      
\usepackage{xcolor}         

\usepackage{caption}        
\usepackage{subcaption}     
\usepackage{mathtools}      

\usepackage{amssymb}        
\usepackage{mathrsfs}       

\DeclarePairedDelimiter{\ceil}{\lceil}{\rceil}
\DeclarePairedDelimiter{\floor}{\lfloor}{\rfloor}
\newcommand{\centroid}[1]{\mathscr{C}({#1})}

\newcommand{\repourl}{https://anonymous.4open.science/r/grokking_explained-2822}

\icmltitlerunning{Grokking Explained: --- A Statistical Phenomenon}

\begin{document}

\twocolumn[
\icmltitle{Grokking Explained: \\ A Statistical Phenomenon}



\icmlsetsymbol{equal}{*}

\begin{icmlauthorlist}
\icmlauthor{Breno W. Carvalho}{ibm,ufrgs}
\icmlauthor{Artur S. d'Avila Garcez}{cul}
\icmlauthor{Lu{\'{\i}}s C. Lamb}{ufrgs}
\icmlauthor{Em{\'{\i}}lio Vital Brazil}{ibm}
\end{icmlauthorlist}

\icmlaffiliation{ibm}{IBM Research, Rio de Janeiro, Brazil}
\icmlaffiliation{ufrgs}{UFRGS, Porto Alegre, Brazil}
\icmlaffiliation{cul}{Department of Computer Science, City, University of London}

\icmlcorrespondingauthor{Breno W. Carvalho}{brenow@ibm.com}

\icmlkeywords{Machine Learning, ICML}

\vskip 0.3in
]



\printAffiliations{}

\begin{abstract}
\textit{Grokking}, or delayed generalization, is an intriguing learning phenomenon where test set loss decreases sharply only after a model's training set loss has converged.  
This challenges conventional understanding of the training dynamics in deep learning networks.  
In this paper, we formalize and investigate grokking, highlighting that a key factor in its emergence is a distribution shift between training and test data.  
We introduce two synthetic datasets specifically designed to analyze grokking. One dataset examines the impact of limited sampling, and the other investigates transfer learning's role in grokking.  
By inducing distribution shifts through controlled imbalanced sampling of sub-categories, we systematically reproduce the phenomenon, demonstrating that while small-sampling is strongly associated with grokking, it is not its cause. Instead, small-sampling serves as a convenient mechanism for achieving the necessary distribution shift.  
We also show that when classes form an equivariant map, grokking can be explained by the model's ability to learn from similar classes or sub-categories.  
Unlike earlier work suggesting that grokking primarily arises from high regularization and sparse data, we demonstrate that it can also occur with dense data and minimal hyper-parameter tuning.  
Our findings deepen the understanding of grokking and pave the way for developing better stopping criteria in future training processes.  
\end{abstract}


\section{Introduction}
\label{intro}
The grokking phenomenon --- a sudden improvement in model performance on unseen data after a period of seemingly stagnant learning, also known as late or delayed generalization --- was first detected in 2021, in the work of \cite{Grokking}. The subsequent literature attributed the phenomenon mostly to three main factors: data sparsity, large initialization values, and high regularization rates. Although the combination of these factors may indeed cause grokking in some situations, they cannot be said to be the cause or to provide a complete explanation of grokking. 

We posit that data sparsity induces grokking by causing a distribution shift during training. To systematically investigate this hypothesis, we construct two synthetic datasets where distribution shifts between training and testing data can be precisely controlled to reproduce grokking, even in the presence of dense data. In the first dataset, classes are subdivided into equidistant subclasses. We show that sub-sampling a subclass can induce grokking with minimal hyperparameter tuning, whereas removing a subclass entirely prevents grokking. This suggests that a weak training signal in data organized into classes and subclasses can compensate for sparsity, enabling late generalization by leveraging relationships among subclasses. These findings highlight the importance of organizing data based on relational structures \cite{kersting}, such as hierarchical class-subclass relationships, and may provide an explanation for the success of geometric and graph-based methods \cite{bronstein} and the broader adoption of neuro-symbolic approaches \cite{Garcez2009}.

In the second dataset, classes form an equivariant map, where transitions between subclasses and classes exhibit a consistent relational structure. Here, we demonstrate that the intensity of grokking depends directly on the distances between classes as defined by subclass relationships. In other words, the relational structure within the data plays a critical role in facilitating late generalization beyond the training data.

In addition to the synthetic datasets, we conduct experiments on the MNIST dataset to explore grokking under a real-world scenario of induced distribution shifts. To create these shifts, we apply a clustering algorithm to each digit based on its representation in a learned feature space. Specifically, we train a ResNet classifier on MNIST and use its latent representations as input for clustering. This approach leverages the ResNet's ability to capture high-level semantic features of the digits, ensuring that the induced clusters are meaningful and aligned with the model's learned representations. The resulting training dataset contains the same digits as the test set but with a different distribution in their representations. These experiments validate the role of distribution shifts in grokking and extend our findings beyond synthetic data.

\begin{figure*}[t]
    \centering

    \begin{subfigure}[t]{0.475\textwidth}
         \centering
         \includegraphics[width=0.3\textwidth]{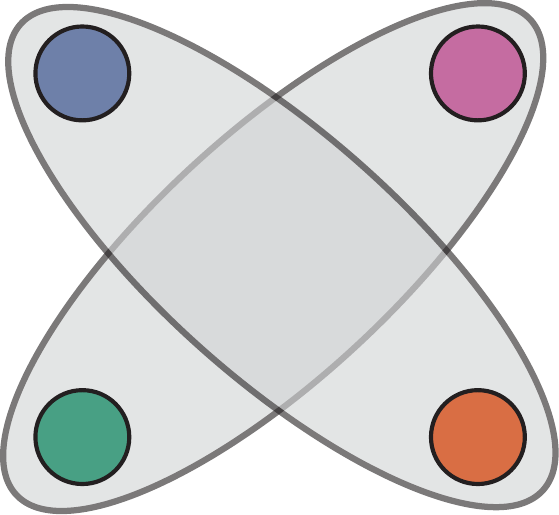}
         \caption{The equidistant subclass dataset, where each class consists of subclasses that are all equidistant from one another. In this illustration, blue and orange are subclasses of one class, while green and purple are subclasses of another. Since four equidistant circles cannot be perfectly represented on a 2D plane, this is a simplified schematic.}
         \label{fig:group_eqdist}
    \end{subfigure}
    \hfill
    \begin{subfigure}[t]{0.475\textwidth}
         \centering
         \includegraphics[width=0.4\textwidth]{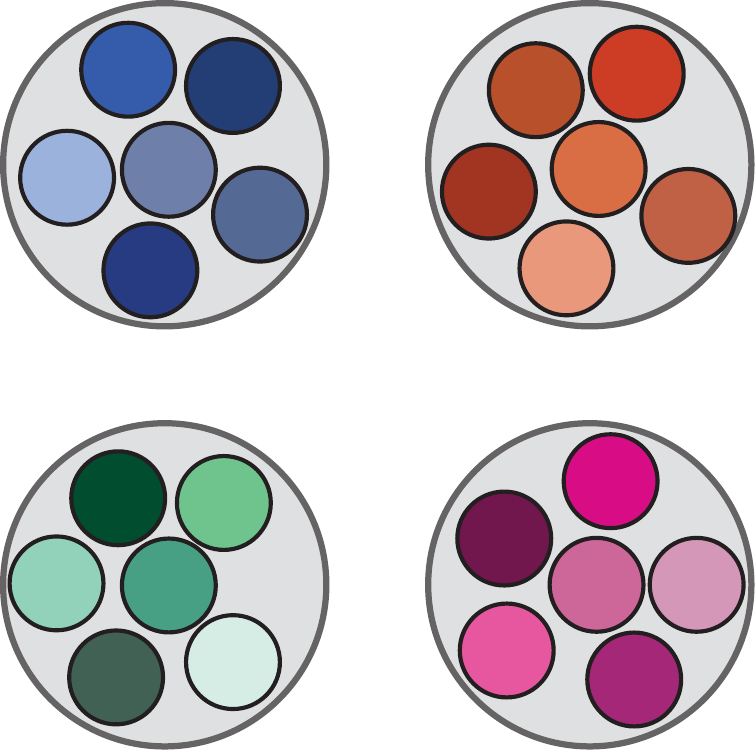}
         \caption{The equivariant subclass dataset, where subclasses are grouped based on similarity. This illustration depicts four classes, each with six subclasses. While the equidistance between subclasses cannot be perfectly visualized in 2D, the schematic emphasizes their relational structure.}
         \label{fig:group_eqvariant}
    \end{subfigure}
    \caption{Schematic representation of the two synthetic datasets used in this study. The equidistant subclass dataset (a) and the equivariant subclass dataset (b).
    }
    \label{fig:synthetic_dataset}
\end{figure*}

Understanding intriguing learning phenomena such as grokking can provide valuable insights into the inner workings of deep neural networks. Late generalization challenges traditional stopping criteria, which rely on monitoring training and test set losses. While it remains difficult to predict when late generalization might occur, this paper takes an important step forward by elucidating how distribution shifts contribute to grokking. Specifically, we explore how modeling class relationships can increase the probability of generalization and guide the design of alternative stopping criteria. These findings are expected to help identify conditions under which models are more likely to generalize after training convergence.

Our study reveals that a specific type of distribution shift, caused by the sampling of data organized into classes and subclasses, can reliably produce late generalization. Unlike prior research, where grokking was observed sporadically, we demonstrate that grokking occurs systematically when certain subcategories or subpatterns in the data are under-represented. Our experiments suggest that a model’s ability to generalize to unseen data depends on subtle data interactions, which may emerge under weak training signals, leading to significant improvements in test performance --- even when performance on specific subcategories remains poor.
The contributions of this paper are as follows:
\begin{itemize} 
\item We introduce two synthetic datasets based on class hierarchies, designed to serve as benchmarks for analyzing late generalization and grokking intensity; 
\item We replicate previous experiments and conduct new analyses, showing that grokking is more strongly associated with distribution shifts than with sparsity and regularization, as previously believed (\url{\repourl} scripts are provided for replication);
\item We perform experiments on the MNIST dataset, where we induce distribution shifts through clustering of digit representations in the latent space of a trained ResNet classifier, validating the generalizability of our findings beyond synthetic data;
\item We report results using both multilayer perceptrons and transformers, concluding that late generalization is largely architecture-independent. 
\end{itemize}

In summary, this paper advances the understanding of grokking as predominantly a distribution shift phenomenon. Based on our findings, we argue that new stopping criteria should be explored, incorporating measures of the likelihood of late generalization and leveraging relational domain knowledge about class hierarchies. These results pave the way for future approaches that embrace grokking as a viable alternative to early stopping.

The remainder of the paper is organized as follows. Section~\ref{sec:background} revisits the related literature and key concepts. Section~\ref{sec:grokking} defines grokking and explores its relationship with distribution shifts. Section~\ref{sec:artificial-datasets} introduces the proposed synthetic datasets and their characteristics for grokking analysis. Section~\ref{sec:results} presents experimental results, demonstrating how grokking can be systematically induced in class-hierarchical data and examining the relationship between grokking intensity and sampling from equivariant subclasses. Finally, Section~\ref{sec:final_remarks} concludes the paper and outlines directions for future research.

\section{Background and Related Work}
\label{sec:background}
Grokking was detected in a few different datasets and different kinds of model architecture. To the best of our knowledge, no definition for grokking exists in the literature, only descriptions and examples of the phenomenon.
Thus, we compiled those descriptions into Definition~\ref{grokking-definition} below, as also illustrated in Figure~\ref{fig:grokking-def}. In the paper that first reported the phenomenon \cite{Grokking}, the grokking signature has a large $\Delta_S$ (see Figure~\ref{fig:grokking-def}) as evaluated predominantly on algorithmic datasets, that is, where the labels are produced by a known algorithm, such as a mathematical operation. How large $\Delta_S$ gets depends on how much the task relies on learning representations, as illustrated by the authors when comparing the latent space of the model before and after grokking takes place. 

In \cite{liu2022Grokking}, a rich study is presented on how different parameter choices may influence the occurrence of grokking. This is done by training a multilayer perceptron (MLP) on MNIST and reducing the size of the training set to 1,000 examples; \cite{liu2022Grokking} do not study what characteristics of the training data might contribute to or avoid grokking. By contrast, our main focus in this paper is to study the training process conditions that create grokking using arbitrary datasets. Nonetheless, \cite{liu2022Grokking} discuss interesting notions of knowledge representation which align with our idea that relational knowledge and class hierarchy is important for grokking.     

In \citet{nanda2023GrokkingProgress}, the focus is on understanding the grokking phenomenon by reverse-engineering neural networks, specifically one-layer transformers trained on modular arithmetic tasks. 
The authors provide interesting insight into how these networks use discrete Fourier transforms and trigonometric identities to achieve emergent behavior, particularly during the grokking phase. They suggest that by defining progress measures for each network component, one should be able to track and understand the training dynamics better. 
Differently from \citet{nanda2023GrokkingProgress}, our work investigates grokking from the perspective of data distribution shifts and statistical properties in general, either for MLPs or Transformers, rather than in the context of a specific network architecture. 

\citet{vzunkovivc2022grokkingPhase} investigate the grokking phenomenon as a phase transition in the context of rule learning. They claim that grokking is a result of the locality of a teacher model and seek to estimate the probability of grokking as being the fraction of sampled attention tensors in a tensor-network map that lead to a linearly separable feature space for a choice of rule. In our work, grokking is formulated purely from a statistical perspective without the need for rule learning. 
\citet{levi2023grokkingLinear} study the occurrence of grokking within linear models through gradient descent. The paper offers an insight that motivated our work: grokking can happen due to data distribution rather than model complexity. The tools used in \cite{levi2023grokkingLinear}, such as the analysis of training and generalization loss dynamics, could be adapted to our work. The paper supports our hypothesis that grokking is caused by distribution shifts in the specific case of liner models. It is generalized in this paper to more complex models. 
\citet{kumar2024grokking} proposes that grokking is a transition from lazy to rich training, that is, when a network proceeds from attempting to learn a linear model from the features that are provided to learn more abstract, derived features. The paper focuses on a kernel regression task, although it reports results on how to try and control grokking on MNIST using an MLP and on modular arithmetic using a one-layer transformer. The authors also claim that weight decay cannot explain grokking. While \citet{kumar2024grokking} places an emphasis on specific model learning, we show that grokking is generally model-independent through our experiments using an MLP and a two-layer transformer (see Appendix~\ref{app:dataset-creation}).



\paragraph{Distribution shift:} A distribution shift, as discussed in \citet{quinonero2022datashift}, refers to a significant change in the statistical properties of a dataset which can lead to poor performance of machine learning models. In this context, a distribution shift typically occurs when the properties of the data used for training differ significantly from the properties of the data encountered during deployment. This discrepancy can result from various factors, such as changes in the data source, user behavior or environment conditions.

\citet{Rabanser2019detect_shift} highlights the effectiveness of two-sample testing methods, particularly those leveraging pre-trained classifiers for dimensionality reduction, in identifying shifts. Additionally, domain-discriminating methods are noted for their ability to qualitatively characterize shifts and assess their potential impact. Since grokking can be viewed as deriving from distribution shifts between training and test sets, the methods and insights from \cite{Rabanser2019detect_shift} could be applied to detect and analyze these shifts in the context of grokking. 

\section{A Definition for Late Generalization (Grokking)}\label{sec:grokking}

\begin{figure}[t]
    \centering
    \includegraphics[width=0.8\linewidth]{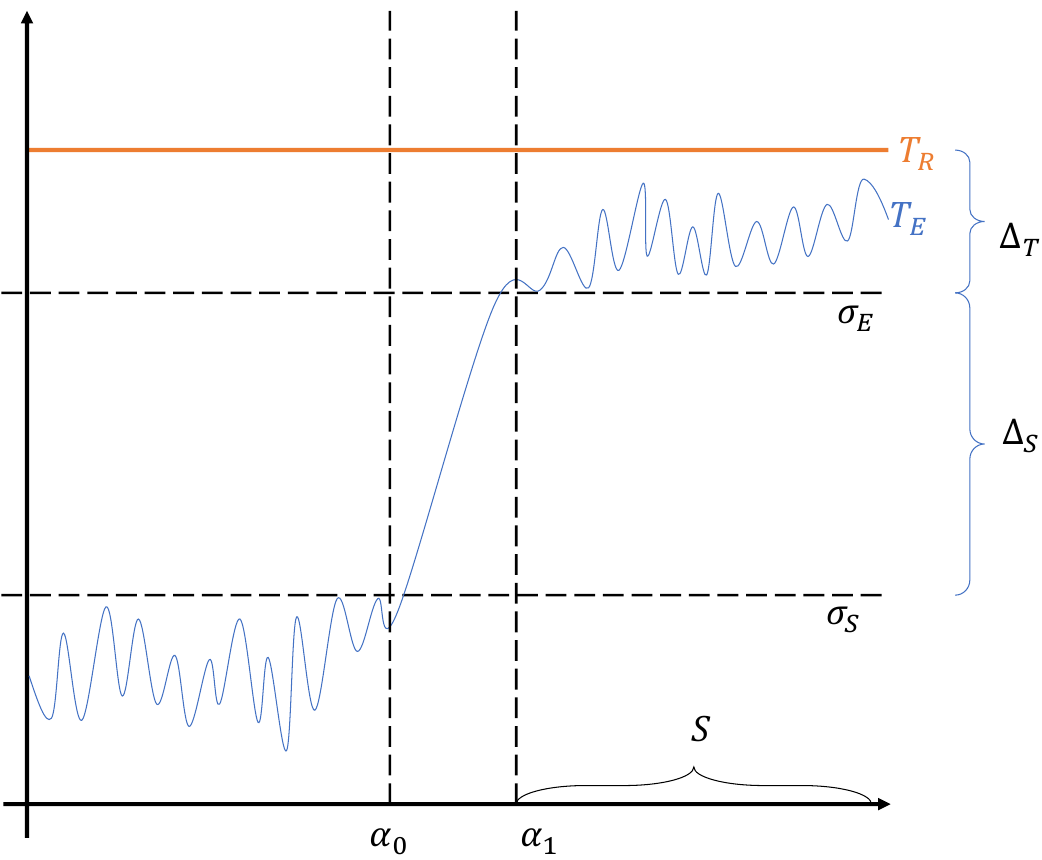}
    \caption{Grokking example: assume that the training set performance has converged ($T_R$, orange curve). There is an inflection point $\alpha_0$ where the model's test set performance ($T_E$, blue curve) escapes a low performing region. The interval between $\alpha_0$ and $\alpha_1$ is the transition phase during which the test set performance transitions to a region much closer to that of the training set performance. By $\Delta_S$, we denote the difference between the upper limit of the low performing region $\sigma_S$ and the lower limit of the high performing region $\sigma_E$. $\Delta_S$ is used to indicate \textit{how much grokking} happened in this training process. By $\Delta_T$, we denote the difference between the limit value of $T_R$ and $\sigma_E$. $\Delta_T$ can be used to indicate how much \textit{room for improvement} there is in the learning process. The interval $S$ is the \textit{support interval}. The longer this interval is without the value of $T_E$ dropping below $\sigma_E$, the more stable the grokking phenomenon.}
    \label{fig:grokking-def}
\end{figure}

As illustrated in Figure \ref{fig:grokking-def}, grokking can be characterized by the transitioning from a low-performing region to a high-performing region as measured on the test set (curve $T_E$ in Fig.\ref{fig:grokking-def}) after the training set performance has converged ($T_R$ in Fig.\ref{fig:grokking-def}). The interval between the inflection points $\alpha_0$ (end of memorization phase) and $\alpha_1$ (start of generalization phase) signifies the transition phase, during which the test set performance gradually improves and aligns with the performance on the training set. 

\begin{definition}
[Grokking]\label{grokking-definition}    
Let $\Delta_T$ denote an interval with lower bound $\sigma_E$ containing the limit value of a converging training set performance curve at time $\alpha_0$. Let $\sigma_S \ll \sigma_E$ denote an upper bound on the test set performance measured up to time point $\alpha_0$. We say that \textit{grokking} took place if at time point $\alpha_1$ > $\alpha_0$, the test set performance is within $\Delta_T$ and remains within $\Delta_T$ for an interval $S \gg (\alpha_1 - \alpha_0)$.
\end{definition}


When training a model, we typically observe training loss and evaluation (test set) loss. Understanding the relationship between these two metrics is essential for interpreting model behavior. If the training loss approaches zero, it suggests that the model has memorized the training data rather than learned to generalize from it. In such cases, one might expect performance on unseen data, represented by the evaluation loss, to degrade due to over-fitting. 
However, if the evaluation loss decreases sharply at a later time, even after the training loss has approached zero, this indicates a capability to generalize, rather than memorization, given a very weak training signal that needs to be understood. 


Our hypothesis is that grokking is a statistical phenomenon derived from a distribution shift between test data and training data, which can be reproduced systematically and which is amplified, not defined, by high regularization and use of specific parametrization and initialization values.

\paragraph{Distribution shifts inducing grokking:}
The phenomenon of grokking, induced by distribution shifts, can be attributed to several key factors.
Firstly, even a faint, i.e. close to zero, training signal may interact with related data within a class or across nearby classes to provide an incentive for models to maintain decision boundaries.
Additionally, regularization serves as a tool to guide models toward simpler solutions with a sparse weight matrix influencing but not determining the occurrence of grokking.
Research by \cite{Grokking} demonstrated grokking in models that do not rely on an internal latent space to represent data, suggesting that while the specific model architecture may be relevant, it is the underlying data distribution and the compounding of learning that will determine late generalization.

In order to confirm or reject our hypothesis, we created two synthetic datasets that allow for controlling the assumed underlying conditions for grokking, as described next. 


\section{New Synthetic Datasets for Grokking}\label{sec:artificial-datasets}

The two synthetic datasets introduced in this paper enable the induction of grokking with minimal tuning of model parameters, avoiding an extensive search of the parameter space. These datasets enable the creation of a shift between training and test set distributions, allowing for user control over how much information each subclass carries, as detailed below. The datasets are available in the supplementary material accompanying the paper.

The primary goal of the datasets is to be linearly separable when considering all sample dimensions $d$, while also being non-trivial to solve even with the addition of noise or extreme down-sampling. Data can be created with an arbitrary number of dimensions $d$. The datasets are defined by projecting classes onto this $d$-dimensional space using multivariate normal distributions, ensuring control over both class definitions and their distances in the $d$-dimensional space.

We define classes $C_1, \dots, C_n$, where each class $C_i$, $1 \leq i \leq n$, is the union of the samples in the disjoint subclasses $C_{i,1}, \dots, C_{i,m}$. We say that $C_i = \bigcup_{j=1}^{m} C_{i,j}$, and $C_{i,p} \cap C_{i,k} = \emptyset$ for all $k \neq p$.


The above structure of classes and subclasses mimics the organization of some real-world datasets, such as COCO \citep{lin2014COCO}, which is used in image segmentation where objects can be part of a larger scene or other objects. More importantly, it provides a straightforward and effective method to induce distribution shifts on demand by down-sampling specific subclasses while preserving the original proportions of the other classes. We also define the dimensionality, $d$, of the dataset.

In the creation of the datasets, we ensure that each subclass $C_{i,j}$ samples a normal multivariate distribution projected onto a $d$-dimensional space for a given $d$. By doing this, we ensure that the dataset is linearly separable in the $d$-dimensional space. Thus, the datasets require a simple classification task solvable with a few optimization steps, but are not so trivial that noise or missing samples would not change model performance.

We cause arbitrary distribution shifts in the datasets by selecting a specific set of subclasses to contain fewer samples than the other subclasses. Without loss of generality, we create all subclasses with the same number of samples in every dataset that we study, i.e. $|C_{i,j}| = s$, for all $i,j$. For example, to create a dataset $D$ with 4,000 samples and four classes, $C_{1}, C_{2}, C_{3}, C_{4}$, each with two subclasses, $C_{i,1}, C_{i,2}, 1 \leq i \leq 4$, each subclass will have 500 samples. To downsample any given subclass, we define $f \in [0, 1]$ such that if $f = 0$ then every sample of that subclass is removed; if $f = 1$ no change is made to the number of samples in the subclass. Given a choice of value for $f$, let $\mathcal{\gamma}_{s}$ denote the number of subclasses to subsample and $\mathcal{\gamma}_{r}$ the number of remaining subclasses. Then, we sample $s_{s}$ examples from each subclass to subsample, and we sample $s_{r}$ examples from the remaining subclasses, as defined in Equation \ref{eq:num_samples_unbalanced}, where $\gamma_D < |D|$ is the total number of samples we want to keep.

\begin{equation}\label{eq:num_samples_unbalanced}
    s_{s} = \ceil[\bigg]{\frac{f\gamma_D}{f\gamma_s + \gamma_r}}
    \quad \text{and} \quad
    s_{r} = \floor[\bigg]{\frac{\gamma_D}{f\gamma_s + \gamma_r}}
\end{equation}


Returning to our example of four classes with two subclasses each, if $\gamma_D = 2000$, $f = 0.2$ and we want to subsample one subclass per class then $s_{s} = \ceil[\bigg]{\frac{0.2 \cdot 2000}{0.2 \cdot 4 + 4}} = 84$ examples and $s_{r} = \floor[\bigg]{\frac{2000}{0.2 \cdot 4 + 4}} = 416$ examples. 
In what follows, we will subsample the subclasses using different values for $f$ up to the point of removing them from the training set altogether. We will study equidistant subclasses (dataset 1) and equivariant subclasses (dataset 2), as described next.  

\noindent \textbf{Dataset 1 (equidistant subclasses)}\label{sec:eqdist-dataset}: to investigate the impact of shifting the training distribution away from the test set distribution in a systematic way, we created a dataset with 4 classes, 
        $C_1,\dots, C_4$, such that each class $C_i$ is composed of two disjoint subclasses $C_{i,1}$ and $C_{i,2}$, $C_i = C_{i,1} \cup C_{i,2}$,
        $C_{i,1} \cap C_{i,2} = \varnothing$.
All the subclasses (no matter which class they belong to) have equidistant centroids. This means that each subclass is a multivariate normal distribution whose central value is equidistant to the central values of every other distribution. Also, if we consider the centroids of each class they will be equidistant to each other. This concept is illustrated in figure~\ref{fig:group_eqdist}.
Although we can generate an unbound number of samples, we use a test set consisting of 10,000 samples.
More detail about the properties of the dataset can be found in Appendix~\ref{app:equidistant_dataset}.

\noindent \textbf{Dataset 2 (equivariant subclasses)}:\label{sec:similar-dataset}
We created a dataset with 4 classes, each class is composed of 10 subclasses.

All the subclasses of a given class have equidistant centroids, calculated as the mean point in the $d$-dimensional space of the samples. The centroids of subclasses from different classes are more distant from each other than they are from the centroids of subclasses of the same class. This means that if $\centroid{C_{i,j}}$ is the centroid of subclass $j$ of class $i$, then:
\[
|\centroid{C_{i,j}} - \centroid{C_{i,l}}| > |\centroid{C_{i,j}} - \centroid{C_{k,l}}|
, i \neq k.
\]

As mentioned previously, with respect to the equidistant dataset, we utilize a test set consisting of 10,000 samples, even though we could generate an unlimited number of samples. This same dataset is employed to examine whether the absence of samples from specific subclasses can be compensated for by the similarity of these subclasses to other classes within the dataset, thereby providing sufficient information for us to test whether grokking takes place as a result of this characteristic.


Further information about the datasets can be found in Appendix~\ref{app:dataset-creation}.

\section{Experimental Results}\label{sec:results}

All our experiments were designed to run on a single GPU, although we ran multiple experiments in parallel to expedite the data analysis process. Each compute node used in the experiments was equipped with an NVIDIA V100 GPU and 16 GB of memory. We conducted experiments in batches, varying the sample size, fraction $f$, and model parameters such as the learning rate and weight decay. Each batch took less than 24 hours to run under these settings. For robustness sake, the results presented in this paper are aggregated from 10 independent runs. 

\subsection{Network Architectures Evaluated}\label{sec:archtectures}


We conducted our experiments using two different architectures: a Multi-layer Perceptron (MLP) and a 2-layer Transformer.
Throughout this paper, we primarily present results for the MLP architecture since the results obtained with both architectures exhibited very similar patterns.
For completeness, the results obtained with the 2-layer Transformer can be found in Appendix~\ref{app:dataset-creation}.
We also provide the code for replicating the experiments in the repository \url{\repourl}.

%
In the experiments reported in this paper we set the learning rate of the models to $10^{-3}$, the weight decay to $10^{-4}$, and have an initialization scale factor of 8 (i.e., we multiply all the weights of the model to this number as described in \citet{liu2022Grokking}, for both architectures tested.

\subsection{Distribution-shift impact on equidistant subclasses}
\begin{figure*}[!ht]
    \centering
    \begin{subfigure}[b]{\textwidth}
         \centering
         \includegraphics[width=4cm]{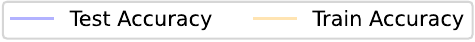}
         
         \includegraphics[width=\textwidth]{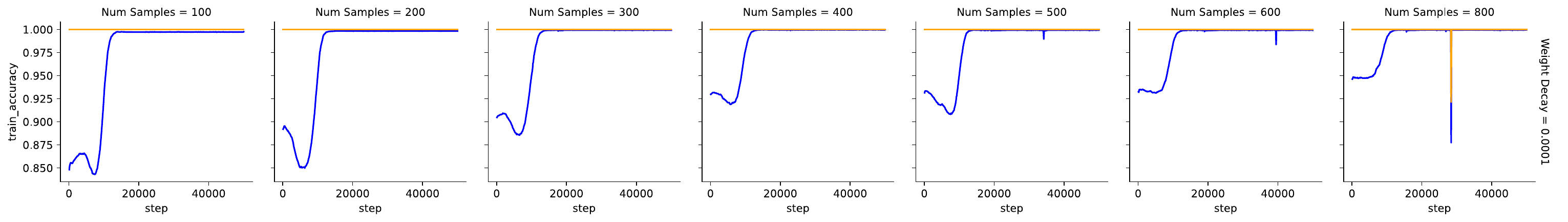}
         \caption{The plots show the effect of varying only the size of the dataset. Grokking is not dependent on the size of the data. It occurs in every case. Grokking intensity, the size of $\Delta_s$, increases as the size of the dataset decreases.
         }
         \label{fig:balanced_data_varying_sample_size}
    \end{subfigure}

    \begin{subfigure}[t]{0.475\linewidth}
         \centering
         \includegraphics[width=\textwidth]{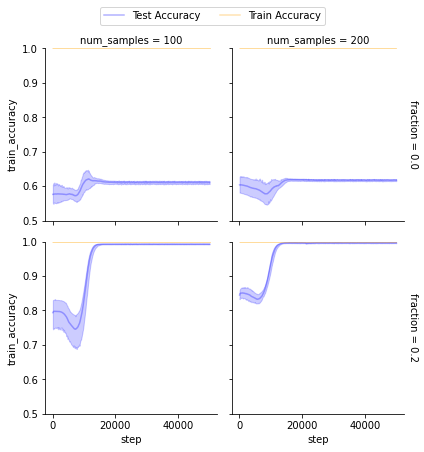}
         \caption{
         Removing a subclass entirely, i.e. $f = 0$, (top plots) prevents grokking, while adding back 20\% of subclass examples (bottom plots) enables grokking within the same maximum number of epochs and same number of samples. Grokking is induced \textit{everything else being equal}.
         }
         \label{fig:small_unbalance_balance_eq}
    \end{subfigure}
    \hfill
    \begin{subfigure}[t]{0.475\linewidth}
         \centering
         \includegraphics[width=\textwidth]{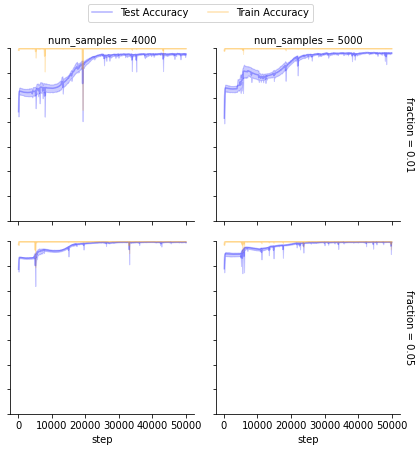}
         \caption{
         Extreme distribution shifts in larger datasets (4000 and 5000 samples) still exhibit grokking, contradicting previous assumptions that grokking only occurs in small datasets.
         It is an example that grokking can occur within larger datasets.
         }
         \label{fig:medium_unbalance_balance}
    \end{subfigure}

    \caption{Average test set curves after the training loss has converged to a very high training set accuracy in the equidistant dataset 1. We examine the effect of small sample sizes (\ref{fig:balanced_data_varying_sample_size}), the effect of an extreme data shift against keeping a few samples from the original distribution (\ref{fig:small_unbalance_balance_eq}), and finally the occurrence of grokking with larger sample sizes, which was assumed not to take place until now (\ref{fig:medium_unbalance_balance}).}
    \label{fig:exp_samples_100-200_frac_0.0-0.2-1.0}
\end{figure*}

\paragraph{Distribution-shift caused by data sparsity:} When performing a balanced down-sampling of our training set, i.e. without forcing any subclass to be sampled more than any other, we find that grokking diminishes as the number of samples increases, only existing in considerably small sample sizes, as depicted in Figure~\ref{fig:balanced_data_varying_sample_size}. This is consistent with the literature, as discussed in section~\ref{sec:grokking}.

\paragraph{Distribution-shift caused by unbalanced equidistant subclass sampling:}
In order to evaluate the impact of distribution shifts in the  occurrence of grokking, we evaluated different data unbalancing created by the use of different values for $f$ as described in Section~\ref{sec:artificial-datasets}. It is clear from Figure~\ref{fig:small_unbalance_balance_eq} that removing completely a subclass has a detrimental effect on generalization, with the models being stuck on a sub-optimal plateau. Nevertheless, adding even a very small number of examples in that subclass (as defined by the fraction e.g. $f = 0.01$ of the missing subclasses) not only allowed the model to generalize well, but allowed the model to generalize late (i.e. grokking occurred). Since this very small fraction of the examples was not visibly relevant for training (was not misclassified) their impact on the test set performance was seen only after training converged and weight decay led the network to a more sparse internal representation.

When we increase the training data size, as mentioned, we see a diminishing effect on grokking. In order to keep analysing this, we then evaluated more aggressive imbalance factors, such as $f = 0.01$ and $f = 0.05$. In a nutshell, Figure~\ref{fig:medium_unbalance_balance} shows that the data distribution shift alone is able to induce late generalization.

\subsection{Distribution-shift impact on equivariant subclasses}
\begin{figure}[t]
    \centering
    \begin{subfigure}[b]{\linewidth}
         \centering
         \includegraphics[width=.8\textwidth]{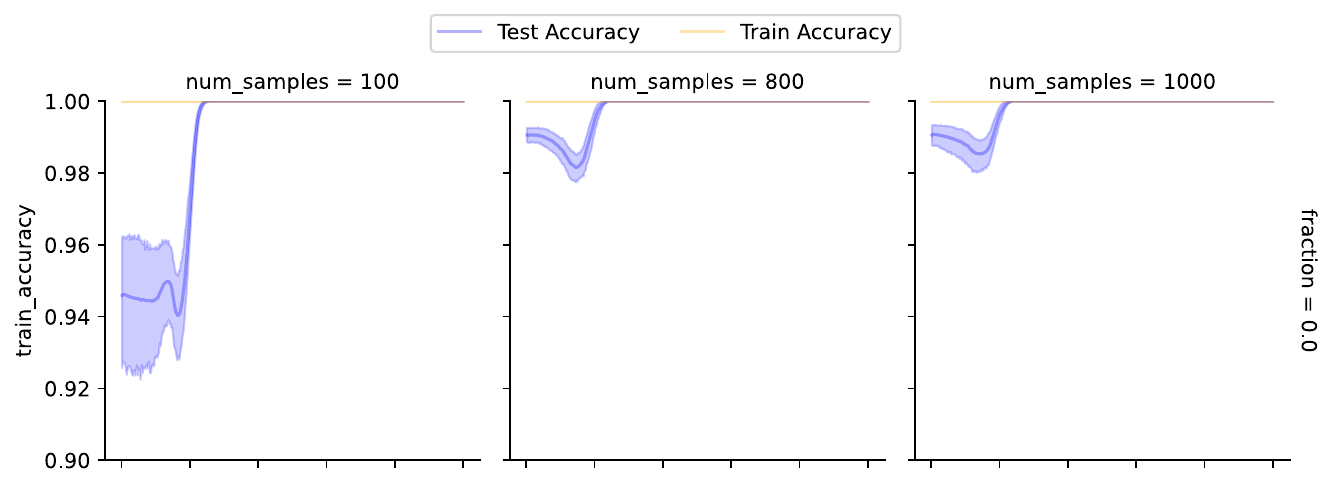}
         \caption{No subclass samples, $f = 0$. Surprisingly, we detect grokking even in the absence of any subclass examples.}
         \label{fig:equivariant-no-sample}
    \end{subfigure}
    
    \vspace{0.5cm}
    
    \begin{subfigure}[b]{\linewidth}
         \centering
         \includegraphics[width=.8\textwidth]{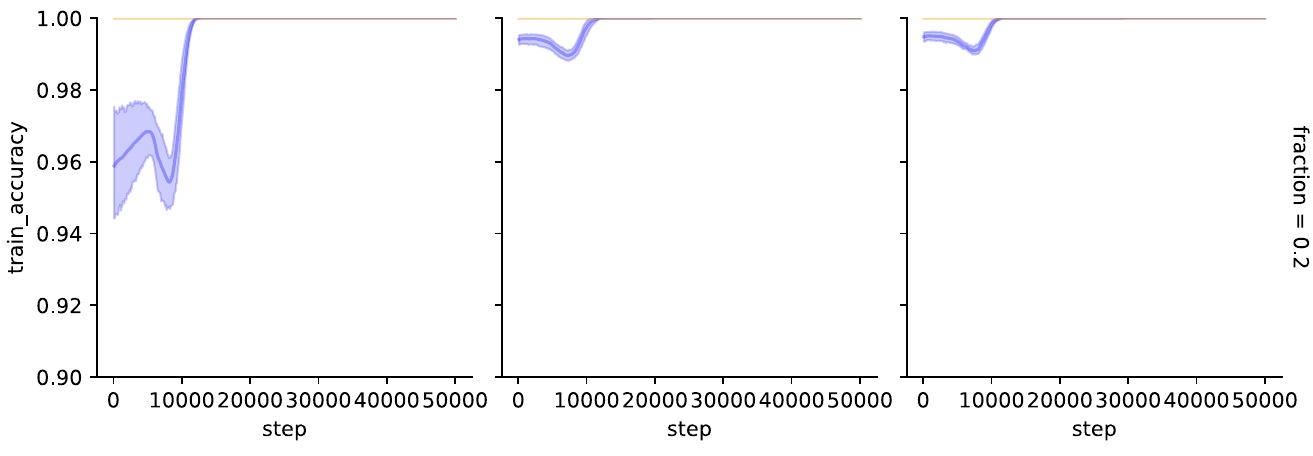}
         \caption{Unbalanced, $f = 0.2$. We still see the grokking pattern, although it already starts at a very high accuracy.}
         \label{fig:equivariant-few-samples}
    \end{subfigure}
    \caption{Investigation of grokking in the equivariant dataset 2. Even without any samples from specific subclasses (Figure~\ref{fig:equivariant-no-sample}), this time the model was able to generalize (differently from the results shown in Fig.~\ref{fig:small_unbalance_balance_eq} for the equidistant dataset 1). This must be due to the similarity, i.e. the proximity in representation space, between the subclasses of a class in dataset 2. In Figure~\ref{fig:equivariant-few-samples}, we see the expected grokking behavior take place both due to the very small sample size that shifts the distribution and the corresponding subclass imbalance. As before, a small sample size increases grokking intensity, but grokking is still present with a larger sample size.}
    \label{fig:shift-equifariant}
\end{figure}

We investigate the effects of sampling strategies on the grokking phenomenon by examining unbalanced sampling of classes. Firstly, we analyze balanced sampling of classes, as illustrated in Figure~\ref{fig:shift-equifariant}. Next, we explore the impact of unbalanced sampling, depicted in Figure~\ref{fig:equivariant-few-samples}. Our findings reveal some form of information leakage among subclasses due to their proximity, which is evidenced by the comparison between the similar graphs with and without samples (Figures~\ref{fig:equivariant-no-sample} and \ref{fig:equivariant-few-samples}).

\paragraph{Distribution-shift caused by unbalanced equivariant subclass sampling:}
The other setting evaluated in our experiments consisted of inducing different imbalances, using different values of $f$ as described in Section~\ref{fig:equivariant-no-sample}. The first noteworthy result here is that we were able to detect grokking even in the cases where samples were removed completely from specific subclasses, as depicted in Figure~\ref{fig:equivariant-few-samples}. This surprising behavior contrasts with the results obtained for the equidistant dataset. It is attributed to the fact that in the equivariant case, subclasses of a given class are closer to each other than to subclasses of other classes, thus carrying information that may result in grokking.

\section{Final remarks}\label{sec:final_remarks}

    This paper has contributed a different perspective to the understanding of the grokking (or late generalization) phenomenon in machine learning. By providing a definition for grokking along with two synthetic datasets that were created to serve as a benchmark for the study of grokking, we have been able to investigate the phenomenon from a new angle, that of data distribution shifts taking place in the context of relations among classes and subclasses of data. We claim that this new perspective on late generalization should offer a better understanding of the phenomenon in connection with relational learning and neuro-symbolic Artificial Intelligence using data and knowledge. 
    Moreover, our observations of emergent behavior when training with very limited examples of a pattern that is nevertheless structured into classes and subclasses underscore the complexity and adaptability of neural networks at learning, the possibility of more efficient learning in future from fewer data and relational knowledge, and the need for further study of learning dynamics in neural networks with implications to training and generalization strategies such as early stopping.

    \paragraph{Broader impact:} Firstly, by providing a deeper understanding of grokking and late generalization, our findings can help researchers and practitioners develop models that generalize better to unseen data. Improved training strategies, informed by insights from how distribution shifts and sampling may affect the probability of grokking, may be created in future. The detailed analysis of grokking available as a Python notebook along with the benchmark datasets may serve as an useful resource for the research community and students new to the field to help them understand complex training dynamics and generalization phenomena. Finally, the insights gained should be applicable across various domains of AI application.

    \paragraph{Limitations:}
    Despite the better characterization of the grokking phenomenon provided in this paper showing that it may not be simply an obscure and rare phenomenon confined to the case of very limited training samples, we have not been able to identify how long a training process may need to continue until late generalization may occur. In practice, while our findings show that grokking is strongly related to training and test set distribution shifts, training strategies are subject to the constraints imposed by real-world dataset characteristics and model parameters.     
    

   \paragraph{Future Steps:} As future work, we plan to inspect the changes taking place in the weight distribution as the network transitions from $\alpha_0$ memorization to late generalization ($\alpha_1$ in Figure~\ref{fig:grokking-def}). In this way, we expect to contribute to the understanding of the underlying mechanism from fitting training data to generalization. Additionally, we aim to connect this work with existing literature \citep{NEURIPS2023_adc98a26, Grokking, liu2022Grokking,nanda2023GrokkingProgress} on how model weights change during phase transitions and to devise entropy metrics for the internal structure of the latent space. Developing these entropy metrics will allow us to quantify the complexity and organization of the latent representations. By focusing on scale and subjective evaluation, some recent developments in deep learning have left behind the main goal of studying generalization via statistically-sound estimation and experimentation. The proper validation of neural networks' generalization capability is of great importance to the research community. This paper is a call for sound generalization estimation to remain a fundamental process of machine learning research. 





\section*{Impact Statement}
This paper presents work whose goal is to advance the field of 
Machine Learning. There are many potential societal consequences 
of our work, none which we feel must be specifically highlighted here.

\bibliography{grokking_bib}
\bibliographystyle{icml2025}

\newpage
\appendix
\onecolumn

\newcommand{\D}{\mathcal{D}}

\section{Dataset creation procedure}\label{app:dataset-creation}


We propose a method for constructing datasets where the distance between different classes can be controlled, enabling systematic analysis of grokking dynamics. Each class $C_i$is characterized by a multivariate distribution $\mathcal{D}_i$. I.e., a data point $x \in \mathbb{R}^n$ is assigned to class $C_i$ if it is drawn from the distribution $\D_i$, allowing precise control over the structure and relationships within the data. To construct the set of distributions $\{\mathcal{D}_0, \dots, \mathcal{D}_k\}$, where $k = 2^p - 1$, we follow a well-defined procedure.

To build the set of distributions $\{\D_0, \cdots, \D_{k}\}$, where $k = 2^p-1$, we create a set of $p$ pairs of sub-distributions $(\mathbf{d_i}^0, \mathbf{d_i}^1)$. Each sub-distribution $\mathbf{d_i}^0$ and $\mathbf{d_i}^1$ has the same dimension $r$, so $x \in \mathbb{R}^{p\times r}$.

To define $\D_j$, we transform the index $j$ into its binary representation with $p$ fixed positions, denoted as $\mathbf{B_p}(j)$. For example, $\mathbf{B_4}(5) = 0101$. We then associate the digit in the $i^\text{th}$ position with the sub-distribution indexed by the digit. For instance, when $p=4$, we can create 15 distributions, with $\D_5= \{\mathbf{d}_0^1, \mathbf{d}_1^0, \mathbf{d}_2^1, \mathbf{d}_3^0\}$.

To ensure that the distance between distributions is preserved in the Euclidean space, we generate $d_{i}^{\rho} = \mathcal{N}(\mu_i^\rho ,\,\Sigma_i)$, where $\mu_i^{\rho} \in \mathbb{R}^r$ is the location and $\Sigma_i \in \mathbb{R}^{r\times r}$ is the covariance matrix. The difference between $\mathbf{d}_i^0$ and $\mathbf{d}_i^1$ lies in their locations, since we use the same covariance matrix. For this work, we set $||\mu_i^0-\mu_i^1|| = 1$ for all $i$.

Now, we can define the distance between two classes using their indices in binary representation. For instance, when $p=4$, the distance between $C_1$ and $C_4$ is calculated as follows: $\mathbf{B_4}(1) = 0001$ and $\mathbf{B_4}(5) = 0101$, so $\mathbf{D}(C_1, C_5) = 1$. Similarly, $\mathbf{D}(C_1, C_2) = 2$.

In this work, we constructed a base set of classes with $p=9$, resulting in 512 distinct classes, and $r=13$, which implies that each data point $x \in \mathbb{R}^{117}$. To simulate real-world scenarios, noise and imbalance are systematically introduced. Imbalance is achieved by varying the number of samples drawn from each class, mimicking uneven class distributions often encountered in practice.
    
\subsection{The Equidistant Dataset}\label{app:equidistant_dataset}
To construct the equidistant dataset, we first create a distance graph $\mathcal{G}$, where each node represents a class and each edge is weighted by the distance between the corresponding classes. We then extract the sub-graph consisting of all edges with a weight of 2. 
Within this sub-graph, we identify the largest clique, which corresponds to the largest subset of classes with a pairwise distance of 2. 

\subsection{The Equivariant Dataset}\label{app:equivariant_dataset}
To construct the equivariant dataset, we reuse the same graph $\mathcal{G}$, but this time extract a clique from the sub-graph of edges with a weight of 6. 
This yields a subset of classes $\{63, 240, 323, 396\}$. 
For each of these classes, we identify all classes that are at a distance of 1 from them. 
This results in 4 super-classes, each comprising 10 classes.
Where we have that: 
$$
 C_i \in SC_k \text{ and } C_j \in SC_l \Rightarrow \left\{\begin{matrix}
\mathbf{D}(C_i, C_j) \geq 4\; \text{ if } \; k \neq l, \\
\mathbf{D}(C_i, C_j) \leq 2 \;\text{ if } \; k = l.
\end{matrix}\right.
$$

The datasets are designed to support controlled experiments focusing on distribution shifts, noise tolerance, and sampling strategies. They simulate challenges often observed in real-world tasks, including imbalanced datasets and noisy measurements, and serve as benchmarks for evaluating model robustness and generalization across various conditions.
By making these datasets publicly available, we aim to facilitate further research into grokking and provide a versatile tool for understanding generalization phenomena in deep learning.

\end{document}